\pdfoutput=1
\documentclass{article} 

\usepackage{acl}

\usepackage{amsmath,amsfonts,bm}

\def\eqref#1{equation~\ref{#1}}

\def\1{\bm{1}}

\DeclareMathAlphabet{\mathsfit}{\encodingdefault}{\sfdefault}{m}{sl}
\SetMathAlphabet{\mathsfit}{bold}{\encodingdefault}{\sfdefault}{bx}{n}

\usepackage{hyperref}
\usepackage{url}
\usepackage{xcolor}
\usepackage{graphicx}
\usepackage{xspace,mfirstuc,tabulary}
\usepackage{times}
\usepackage{latexsym}
\usepackage{booktabs}
\usepackage{multirow}
\usepackage{comment}
\usepackage{xspace}
\usepackage{color, colortbl}

\newcommand*{\yoruba}{Yor\`ub\'a\xspace}
\newcommand*{\zulu}{isiZulu\xspace}

\newcommand*\samethanks[1][\value{footnote}]{\footnotemark[#1]}

\definecolor{Color}{gray}{0.9}

\title{Adapting Pre-trained Language Models to African Languages via Multilingual Adaptive Fine-Tuning}

\author{Jesujoba O. Alabi\thanks{* Equal contribution.}, \ David Ifeoluwa Adelani\samethanks, \ Marius Mosbach, and Dietrich Klakow\\
Spoken Language Systems (LSV), Saarland University,
Saarland Informatics Campus, Germany\\
\texttt{\{jalabi,didelani,mmosbach,dklakow\}@lsv.uni-saarland.de}
}

\begin{document}

\maketitle

\begin{abstract}
Multilingual pre-trained language models (PLMs) have demonstrated impressive performance on several downstream tasks for both high-resourced and low-resourced languages.  However, there is still a large performance drop for languages unseen during pre-training, especially African languages. One of the most effective approaches to adapt to a new language is \textit{language adaptive fine-tuning} (LAFT) --- fine-tuning a multilingual PLM on monolingual texts of a language using the pre-training objective. However, 
adapting to a target language individually takes a large disk space and limits the cross-lingual transfer abilities of the resulting models because they have been specialized for a single language. In this paper, 
we perform \textit{multilingual adaptive fine-tuning} on 17 most-resourced African languages and three other high-resource languages widely spoken on the African continent 
to encourage cross-lingual transfer learning. 
To further specialize the multilingual PLM, we removed vocabulary tokens from the embedding layer that corresponds to non-African writing scripts before MAFT, thus reducing the model size by around 50\%. 
Our evaluation on two multilingual PLMs (AfriBERTa and XLM-R) and three NLP tasks (NER, news topic classification, and sentiment classification) shows that our approach is competitive to applying LAFT on individual languages while requiring significantly less disk space. Additionally, we show that our adapted PLM also improves the zero-shot cross-lingual transfer abilities of parameter efficient fine-tuning methods.

\end{abstract}

\section{Introduction}
Recent advances in the development of multilingual pre-trained language models (PLMs) like mBERT~\citep{devlin-etal-2019-bert}, XLM-R~\citep{conneau-etal-2020-unsupervised}, and RemBERT~\citep{chung2021rethinking} have led to significant performance gains on a wide range of cross-lingual transfer tasks. Due to the \textit{curse of multilinguality}~\citep{conneau-etal-2020-unsupervised} --- a trade-off between language coverage and
model capacity --- and non-availability of pre-training corpora for many low-resource languages, multilingual PLMs are often trained on about 100 languages. Despite the limitations of language coverage,  multilingual PLMs have been shown to transfer to several low-resource languages unseen during pre-training. Although, there is still a large performance gap compared to languages seen during pre-training. 

One of the most effective approaches to adapt to a new language is \textit{language adaptive fine-tuning} (LAFT) --- fine-tuning a multilingual PLM on monolingual texts in the target language using the same pre-training objective. This has been shown to lead to big gains on many cross-lingual transfer tasks~\citep{pfeiffer-etal-2020-mad}, and low-resource languages~\citep{muller-etal-2021-unseen,chau-smith-2021-specializing}, including African languages~~\citep{alabi-etal-2020-massive,adelani-etal-2021-masakhaner}. Nevertheless, 
adapting a model to each target language individually takes large disk space, and limits the cross-lingual transfer abilities of the resulting models because they have been specialized to individual languages~\citep{beukman2021analysing}. 

An orthogonal approach to improve the coverage of low-resource languages is to include them in the pre-training data. An example for this approach is AfriBERTa~\citep{ogueji-etal-2021-small}, which was trained from scratch on 11 African languages.  
A downside of this approach is that it is resource intensive in terms of data and compute.  

Another alternative approach is parameter efficient fine-tuning like Adapters ~\citep{pfeiffer-etal-2020-mad} and sparse fine-tuning~\citep{ansell2021composable}, where the model is adapted to new languages by using a sparse network trained on a small monolingual corpus. Similar to LAFT, it requires adaptation for every new target language. Although it takes little disk space, all target language-specific parameters need to be stored. 

In this paper, we propose \textit{multilingual adaptive fine-tuning (MAFT)}, a language adaptation to multiple languages at once. We perform language adaptation on the 17 most-resourced African languages (Afrikaans, Amharic, Hausa, Igbo, Malagasy, Chichewa, Oromo, Naija, Kinyarwanda, Kirundi, Shona, Somali, Sesotho, Swahili, isiXhosa, \yoruba, \zulu) and three other high-resource language widely spoken on the continent (English, French, and Arabic) \textit{simultaneously} to provide a single model for cross-lingual transfer learning for African languages. 
To further specialize the multilingual PLM, we follow the approach of \citet{abdaoui-etal-2020-load} and remove vocabulary tokens from the embedding layer that correspond to 
non-Latin and non-Ge'ez (used by Amharic) scripts
before MAFT, thus effectively reducing the model size by  50\%. 

Our evaluation on two multilingual PLMs (AfriBERTa and XLM-R) and three NLP tasks (NER, news topic classification and sentiment classification) shows that our approach is competitive to performing LAFT on the individual languages, with the benefit of having a single model instead of a separate model for each of the target languages. Also, we show that our adapted PLM improves the zero-shot cross-lingual transfer abilities of parameter efficient fine-tuning methods like Adapters~\cite{pfeiffer-etal-2020-mad} and sparse fine-tuning~\citep{ansell2021composable}. 


As an additional contribution, and in order to cover more diverse African languages in our evaluation, we create a new evaluation corpus, ANTC -- \underline{\textbf{A}}frican  \underline{\textbf{N}}ews \underline{\textbf{T}}opic \underline{\textbf{C}}lassification -- for  
Lingala, Somali, Naija, Malagasy, and isiZulu 
from pre-defined news categories of VOA, BBC, Global Voices, and Isolezwe newspapers. To further the research on NLP for African languages,  we make our code and data publicly available.\footnote{\url{https://github.com/uds-lsv/afro-maft}} Additionally, our models are available via HuggingFace.\footnote{\url{https://huggingface.co/Davlan}} 

\section{Related Work}
\label{related}
\paragraph{Multilingual PLMs for African languages} The success of multilingual PLMs such as mBERT~\citep{devlin-etal-2019-bert} and XLM-R~\citep{conneau-etal-2020-unsupervised} for cross-lingual transfer in many natural language understanding tasks has encouraged the continuous development of multilingual models~\citep{luo-etal-2021-veco,chi-etal-2021-infoxlm,ouyang-etal-2021-ernie,chung2021rethinking,He2021DeBERTaV3ID}. Most of these models cover 50 to 110 languages and only few African languages are represented due to lack of large monolingual corpora on the web. To address this under-representation, regional multilingual PLMs have been trained from scratch such as AfriBERTa~\citep{ogueji-etal-2021-small} or adapted from existing multilingual PLM through LAFT~\citep{alabi-etal-2020-massive,pfeiffer-etal-2020-mad,muller-etal-2021-unseen,adelani-etal-2021-masakhaner}. AfriBERTa is a relatively small multilingual PLM (126M parameters) trained using the RoBERTa architecture and pre-training objective on 11 African languages. 
However, it lacks coverage of languages from the southern region of the African continent, specifically the southern-Bantu languages. 
In our work, we extend to those languages since only a few of them have large ($>$100MB size) monolingual corpus. 
We also do not specialize to a single language but apply MAFT which allows multilingual adaptation and preserves downstream performance on both high-resource and low-resource languages.  

\paragraph{Adaptation of multilingual PLMs} It is not unusual for a new multilingual PLM to be initialized from an existing model. For example, \citet{chi-etal-2021-infoxlm} trained InfoXLM by initializing the weights from XLM-R before training the model on a joint monolingual and translation corpus. Although they make use of a new training objective during adaptation. Similarly, \citet{Tang2020MultilingualTW} extended the languages covered by mBART~\citep{liu-etal-2020-multilingual-denoising} from 25 to 50 by first modifying the vocabulary and initializing the model weights of the original mBART before fine-tuning it on a combination of monolingual texts from the original 25 languages in addition to 25 new languages. Despite increasing the number of languages covered by their model, they did not observe a significant performance drop on downstream tasks. We take inspiration from these works for applying MAFT on African languages, but we do not modify the training objective during adaptation nor increase the vocabulary.

\paragraph{Compressing PLMs} One of the most effective methods for creating smaller PLMs is distillation where a small student model is trained to reproduce the behaviour of a larger teacher model. This has been applied to many English PLMs~\citep{Sanh2019DistilBERTAD,jiao-etal-2020-tinybert,sun-etal-2020-mobilebert,liu-etal-2020-fastbert} and a few multilingual PLMs~\citep{wang_minilm,wang-etal-2021-minilmv2}. However, it often leads to a drop in performance compared to the teacher PLM. An alternative approach that does not lead to a drop in performance has been proposed by \citet{abdaoui-etal-2020-load} for multilingual PLM. They removed unused vocabulary tokens from the embedding layer. This simple method significantly reduces the number of embedding parameters thus reducing the overall model size since the embedding layer contributes the most to the total number of model parameters. In our paper, we combine MAFT with the method proposed by \citet{abdaoui-etal-2020-load} to reduce the overall size of the resulting multilingual PLM for African languages. 
This is crucial especially because people from under-represented communities in Africa may not have access to powerful GPUs in order to fine-tune large PLMs. Also, Google Colab\footnote{\url{https://colab.research.google.com/}} (free-version), which is widely used by individuals from under-represented communities without access to other compute resources, cannot run large models like e.g. XLM-R. 
Hence, it is important to provide smaller models that still achieve competitive downstream performance 
to these communities.

\begin{table*}[t]
\footnotesize
 \begin{center}
 \scalebox{0.85}{
\begin{tabular}{p{30mm}rrrp{60mm}r}
\toprule
 \multirow{2}{*}{\textbf{Domain}} &  \multicolumn{3}{c}{\textbf{Number of sentences}} & \multirow{2}{*}{\textbf{Classes}} &  \multirow{2}{*}{\textbf{Number of classes}}  \\
 \cmidrule{2-4}
  & \textbf{Train} & \textbf{Dev} & \textbf{Test} &   \\
\midrule
\textbf{\emph{Newly created datasets}} \\ 
Lingala (lin) & 1,536 & 220  & 440  & Rdc, Politiki/Politique, Bokengi/Securite, Justice, Bokolongono/Santé/Medecine & 5  \\
Naija (pcm) & 1,165 & 167 & 333 & Entertainment, Africa, Sport, Nigeria, World  & 5 \\
Malagasy (mlg) & 3,905 & 559 & 1,117 & Politika (Politics), Kolontsaina (Culture), Zon'olombelona (Human Rights), Siansa\_sy\_Teknolojia (Science and Technology) ,Tontolo\_iainana (Environment)  & 5 \\
Somali (som) & 10,072 & 1,440 & 2879 & Soomaaliya (Somalia), Wararka (News), Caalamka (World), Maraykanka (United States), Afrika (Africa) & 6 \\
\zulu (zul) & 2,961 & 424 & 847 & Ezemidlalo (Sports), Ezokungcebeleka (Recreation), Imibono (Ideas), Ezezimoto (Automotive), Intandokazi (Favorites) & 5          \\
\midrule
\textbf{\emph{Existing datasets}} \\
Amharic (amh) & 36,029 & 5,147 & 10,294 & Local News, Sport, Politics, International News, Business, Entertainment  & 6          \\
English (eng) & 114,000 & 6,000 & 7,600 & World, Sports, Business, Sci/Tech & 4         \\
Hausa (hau) & 2,045 & 290 & 582  & Africa, World, Health, Nigeria, Politics & 5         \\
Kinyarwanda (kin) & 16,163 & 851 & 4,254 & Politics, Sport, Economy, Health, Entertainment, History, Technology, Tourism, Culture, Fashion, Religion, Environment, Education, Relationship & 14           \\
Kiswahili (swa) & 21,096 & 1,111 & 7,338 & Uchumi (Economic), Kitaifa (National), Michezo (Sports), Kimataifa (International), Burudani (Recreation), Afya (Health) & 6         \\
\yoruba (yor) & 1,340 & 189 & 379 & Nigeria, Africa, World, Entertainment, Health, Sport, Politics & 7         \\
\bottomrule
\end{tabular}
}
\footnotesize
  \caption{Number of sentences in training, development and test splits. We provide automatic translation of some of the African language words to English (in Parenthesis) using Google Translate. }
  \label{tab:dev_test_split}
\end{center}
\end{table*}

\paragraph{Evaluation datasets for African languages}
One of the challenges of developing (multilingual) PLMs for African languages is the lack of evaluation corpora. There have been many efforts by communities like Masakhane to address this issue~\citep{nekoto_etal_2020_participatory,adelani-etal-2021-masakhaner}. We only find two major evaluation benchmark datasets that cover a wide range of African languages: one for named entity recognition (NER)~\citep{adelani-etal-2021-masakhaner} and one for sentiment classification~\citep{Muhammad2022NaijaSentiAN}. In addition, there are also several news topic classification datasets~\citep{hedderich-etal-2020-transfer,niyongabo-etal-2020-kinnews,Azime2021AnAN} but they are only available for a few African languages. Our work contributes novel news topic classification datasets (i.e. ANTC) for additional five African languages: 
Lingala, Naija, Somali, isiZulu, and Malagasy. 

\section{Data}
\label{data}

\subsection{Adaptation corpora}
\label{exist_corpus}
We perform MAFT on 17 African languages
Afrikaans, Amharic, Hausa, Igbo, Malagasy, Chichewa, Oromo, Naija, Kinyarwanda, Kirundi, Shona, Somali, Sesotho, Swahili, isiXhosa, \yoruba, \zulu)
covering the major African language families and 3 high resource languages (Arabic, French, and English) widely spoken in Africa. We selected the African languages based on the availability of a (relatively) large amount of monolingual texts. We obtain the monolingual texts from three major sources: the mT5 pre-training corpus which is based on Common Crawl Corpus\footnote{\url{https://commoncrawl.org/}}~\citep{xue-etal-2021-mt5}
, the British Broadcasting Corporation (BBC) News, Voice of America News\footnote{\url{https://www.voanews.com}}~\citep{DBLP:journals/corr/abs-2201-05609}, and some other news websites based in Africa.  \autoref{tab:monolingual_corpus} in the Appendix provides a summary of the monolingual data, including their sizes and sources. We pre-processed the data by removing lines that consist of numbers or punctuation only, and lines with less than six tokens. 

\subsection{Evaluation tasks}
\label{new_corpus}

We run our experiments on two sentence level classification tasks: news topic classification and sentiment classification, and one token level classification task: NER. We evaluate our models on English as well as diverse African languages with different linguistic characteristics.

\subsubsection{Existing datasets}
\paragraph{NER} For the NER task we evaluate on the MasakhaNER dataset \citep{adelani-etal-2021-masakhaner}, a manually annotated dataset covering 10 African languages (Amharic, Hausa, Igbo, Kinyarwanda, Luganda, Luo, Naija, Kiswahili, Wolof, and Yorùbá) with texts from the news domain. For English, we use data from the CoNLL 2003 NER task~\citep{tjong-kim-sang-de-meulder-2003-introduction} also containing texts from the news domain. For isiXhosa, we use the data from \citet{eiselen-2016-government}. Lastly, to evaluate on Arabic we make use of the ANERCorp dataset~\citep{benajiba_arabic_ner,obeid-etal-2020-camel}.

\paragraph{News topic classification}
We use existing news topic datasets for Amharic~\citep{Azime2021AnAN}, English -- AG News corpus --~\citep{Zhang2015CharacterlevelCN}, Kinyarwanda -- KINNEWS --~\citep{niyongabo-etal-2020-kinnews}, Kiswahili -- new classification dataset--~\citep{davis_david_2020_5514203}, and both \yoruba and Hausa~\citep{hedderich-etal-2020-transfer}. For dataset without a development set, we randomly sample {5\%} of their training instances and use them as a development set.

\paragraph{Sentiment classification} We use the NaijaSenti multilingual Twitter sentiment analysis corpus \citep{Muhammad2022NaijaSentiAN}. This is a large code-mixed and monolingual sentiment analysis dataset, manually annotated  for 4 Nigerian languages: Hausa, Igbo, \yoruba and Pidgin. Additionally, we evaluate on the Amharic, and English Twitter sentiment datasets by \citet{yimam-etal-2020-exploring} and \citet{rosenthal2017semeval}, respectively. For all datasets above, we only make use of tweets with positive, negative and neutral sentiments.

\subsubsection{Newly created dataset: ANTC corpus}

We created a novel dataset, ANTC --- \underline{\textbf{A}}frican  \underline{\textbf{N}}ews \underline{\textbf{T}}opic \underline{\textbf{C}}lassification for five African languages. We obtained data from three different news sources: VOA, BBC\footnote{\url{https://www.bbc.com/pidgin}}, Global Voices\footnote{\url{https://mg.globalvoices.org/}}, and isolezwe\footnote{\url{https://www.isolezwe.co.za}}. From the VOA data we created datasets for Lingala and Somali. 
We obtained the topics from data released by \citet{DBLP:journals/corr/abs-2201-05609} and used the provided URLs to get the news category from the websites. For Naija, Malagasy and \zulu, we scrapped news topic from the respective news website (BBC Pidgin, Global Voices, and isolezwe respectively) directly base on their category. We noticed that some news topics are not mutually exclusive to their categories, therefore, we filtered such topics with multiple labels. Also, we ensured that each category has at least 200 samples. The categories include but are not limited to: Africa, Entertainment, Health, and Politics. 
The pre-processed datasets were divided into training, development, and test sets using stratified sampling with a ratio of 70:10:20. \autoref{tab:dev_test_split} provides details about the dataset size and news topic information.


\section{Pre-trained Language Models}

\label{mlm}
For our experiments, we make use of different multilingual PLMs that have been trained using a masked language model objective on large collections of monolingual texts from several languages. \autoref{tab:plm_languages} shows the number of parameters as well as the African languages covered by each of the models we consider.

\begin{enumerate}
    \item XLM-R~\citep{conneau-etal-2020-unsupervised} has been pre-trained on 100 languages including eight African languages. We make use of both XLM-R-base and XLM-R-large for MAFT with 270M and 550M parameter sizes respectively. Although, for our main experiments, we make use of XLM-R-base.  
    \item AfriBERTa~\citep{ogueji-etal-2021-small} has been pre-trained only on African languages. Despite its smaller parameter size (126M), it has been shown to reach competitive performance to XLM-R-base on African language datasets~\citep{adelani-etal-2021-masakhaner,hedderich-etal-2020-transfer}.
    \item XLM-R-miniLM~\citep{wang_minilm} is a distilled version of XLM-R-large with only 117M parameters. 
    
\end{enumerate}
\begin{table}[t]
 \begin{center}
 \scalebox{0.82}{
 \footnotesize
  \begin{tabular}{p{22mm}rp{48mm}}
    \toprule
    \textbf{PLM} & \textbf{\# Lang.} & \textbf{African languages covered} \\
    \midrule
    XLM-R-base (270M) &100 & \texttt{afr}, \textbf{\texttt{amh}}, \textbf{\texttt{hau}}, \textbf{\texttt{mlg}}, \texttt{orm}, \textbf{\texttt{som}}, \textbf{\texttt{swa}}, \textbf{\texttt{xho}}  \\
    AfriBERTa-large (126M) & 11 &  \textbf{\texttt{amh}}, \textbf{\texttt{hau}}, \textbf{\texttt{ibo}}, \textbf{\texttt{kin}}, \texttt{run}, \texttt{orm}, \textbf{\texttt{pcm}}, \textbf{\texttt{som}}, \textbf{\texttt{swa}}, \texttt{tir}, \textbf{\texttt{yor}}  \\
    XLM-R-miniLM (117M) & 100 & \texttt{afr}, \textbf{\texttt{amh}}, \textbf{\texttt{hau}}, \textbf{\texttt{mlg}}, \texttt{orm}, \textbf{\texttt{som}}, \textbf{\texttt{swa}}, \textbf{\texttt{xho}}  \\
    XLM-R-large (550M) &100 & \texttt{afr}, \textbf{\texttt{amh}}, \textbf{\texttt{hau}}, \textbf{\texttt{mlg}}, \texttt{orm}, \textbf{\texttt{som}}, \textbf{\texttt{swa}}, \textbf{\texttt{xho}}  \\
    AfroXLMR* (117M-550M) & 20 & \texttt{afr}, \textbf{\texttt{amh}}, \textbf{\texttt{hau}}, \textbf{\texttt{ibo}}, \textbf{\texttt{kin}}, \texttt{run} \textbf{\texttt{mlg}}, \texttt{nya}, \texttt{orm}, \textbf{\texttt{pcm}}, \textbf{\texttt{sna}}, \textbf{\texttt{som}}, \texttt{sot}, \textbf{\texttt{swa}}, \textbf{\texttt{xho}}, \textbf{\texttt{yor}}, \textbf{\texttt{zul}}   \\

    \bottomrule
  \end{tabular}
  }
  \vspace{-3mm}
  \caption{Language coverage and size for pre-trained language models. Languages in \textbf{bold} have evaluation datasets for either NER, news topic classification or sentiment analysis.}
  \label{tab:plm_languages}
  \end{center}
\end{table}

\begin{table*}[t]
\footnotesize
 \begin{center}
  \resizebox{\textwidth}{!}{%
  \begin{tabular}{ll|rrrrrrrrrrrrr|r}
    \toprule
    \textbf{Model} &\textbf{Size} &\textbf{\texttt{amh}} &\textbf{\texttt{ara}}  &\textbf{\texttt{eng}} &\textbf{\texttt{hau}}  & \textbf{\texttt{ibo}} & \textbf{\texttt{kin}} & \textbf{\texttt{lug}} & \textbf{\texttt{luo}} & \textbf{\texttt{pcm}} & \textbf{\texttt{swa}} & \textbf{\texttt{wol}} & \textbf{\texttt{xho}} & \textbf{\texttt{yor}} & \textbf{avg} \\
    \midrule
    \multicolumn{4}{l}{\texttt{Finetune}} \\
    XLM-R-miniLM & 117M & 69.5 & 76.1 & 91.5 & 74.5 & 81.9 & 68.6 & 64.7 & 11.7 & 83.2 & 86.3 & 51.7 & 69.3 & 72.0 & 69.3\\
    AfriBERTa & 126M & 73.8 & 51.3 & 89.0 & 90.2 & 87.4 & 73.8 & 78.9 & 70.2 & 85.7 & 88.0 & 61.8 & 67.2 & 81.3 & 76.8 \\
    \rowcolor{Color}
    XLM-R-base & 270M & 70.6  & 77.9 & 92.3 & 89.5 & 84.8 & 73.3 & 79.7 & 74.9 & 87.3 & 87.4 & 63.9 & 69.9 & 78.3 & 79.2 \\
    XLM-R-large & 550M & 76.2  & \textbf{79.7} & 93.1 & 90.5 & 84.1 & 73.8 & 81.6 & 73.6 & 89.0 & 89.4 & 67.9 & 72.4 & 78.9 & 80.8 \\
    \midrule
    \multicolumn{4}{l}{\texttt{MAFT + Finetune}} \\
    XLM-R-miniLM  & 117M& 69.7  & 76.5 & 91.7 & 87.7 & 83.5 & 74.1 & 77.4 & 17.5 & 85.5 & 86.0 & 59.0 & 72.3 & 75.1 & 73.5 \\
    AfriBERTa  & 126M & 72.5  & 40.9 & 90.1 & 89.7 & 87.6 & 75.2 & 80.1 & 69.6 & 86.5 & 87.6 & 62.3 & 71.8 & 77.0 & 76.2\\
    \rowcolor{Color}
    XLM-R-base  & 270M& 76.1  & \textbf{79.7} & \textbf{92.8} & 91.2 & 87.4 & \textbf{78.0} & 82.9 & 75.1 & 89.6 & 88.6 & 67.4 & 71.9 & 82.1 & 81.8\\
    XLM-R-base-v70k  & 140M & 70.1  & 76.4 & 91.0 & 91.4 & 86.6 & 77.5 & 83.2 & \textbf{75.4} & 89.0 & 88.7 & 65.9 & 72.4 & 81.3 & 80.7\\
    \midrule
    \rowcolor{Color}
    XLM-R-base+LAFT  & 270M x 13 & \textbf{78.0}  & 79.1  & 91.3 & \textbf{91.5} & \textbf{87.7} & 77.8 & \textbf{84.7} & 75.3 & \textbf{90.0} & \textbf{89.5} & \textbf{68.3} & \textbf{73.2} & \textbf{83.7} & \textbf{82.3}\\
    \bottomrule
  \end{tabular}
  }
    \caption{\textbf{NER} model comparison, showing F1-score on the test sets after 50 epochs averaged over 5 runs. Results are for all 4 tags in the dataset: PER, ORG, LOC, DATE/MISC. For LAFT, we multiplied the size of XLM-R-base by the number of languages as LAFT results in a single model per language.}
  \label{tab:ner}
  \end{center}
  
\end{table*}

\paragraph{Hyper-parameters for baseline models} We fine-tune the baseline models for NER, news topic classification and sentiment classification for 50, 25, and 20 epochs respectively. We use a learning rate of 5e-5 for all the task, except for sentiment classification where we use 2e-5 for XLM-R-base and XLM-R-large. The maximum sequence length is 164 for NER, 500 for news topic classification, and 128 for sentiment classification. The adapted models also make use of similar hyper-parameters.

\section{Multilingual Adaptive Fine-tuning}
\label{sec:maft_sec}
We introduce MAFT as an approach to adapt a multi-lingual PLM to a new set of languages. Adapting 
PLMs has been shown to be effective when adapting to a new domain~\citep{gururangan-etal-2020-dont} or language \citep{pfeiffer-etal-2020-mad,alabi-etal-2020-massive,muller-etal-2021-unseen,adelani-etal-2021-masakhaner}. While previous work on multilingual adaptation has mostly focused on autoregressive sequence-to-sequence models such as mBART~\citep{Tang2020MultilingualTW},  
in this work, we adapt 
non-autoregressive masked PLMs on monolingual corpora covering 20 languages. Crucially, during adaptation we use the same objective that was also used during pre-training. 
The models resulting from MAFT can then be fine-tuned on supervised NLP downstream tasks. 
We name the model resulting after applying MAFT to XLM-R-base and XLM-R-miniLM as \textit{AfroXLMR-base} and \textit{AfroXLMR-mini}, respectively. 
For adaptation, we train on a combination of the monolingual corpora used for AfriMT5 adaptation by \citet{adelani-etal-2022-thousand}. Details for each of the monolingual corpora and languages are provided in Appendix \ref{appendix:maft-corpora}.

\paragraph{Hyper-parameters for MAFT}
The PLMs were trained for 3 epochs with a learning rate of 5e-5 using huggingface transformers \citep{wolf-etal-2020-transformers}. We use of a batch size of 32 for AfriBERTa and a batch size 10 for the other PLMs. 

\subsection{Vocabulary reduction}
\label{reduce}
Multilingual PLMs come with various parameter sizes, the larger ones having more than hundred million parameters, which makes fine-tuning and deploying such models a challenge due to resource constraints. 
One of the major factors that contributes to the parameter size of these models is the embedding matrix whose size is a function of the vocabulary size of the model. While a large vocabulary size is essential for a multilingual PLM trained on hundreds of languages, some of the tokens in the vocabulary can be removed when they are irrelevant to the domain or language considered in the downstream task, thus reducing the vocabulary size of the model. Inspired by \citet{abdaoui-etal-2020-load}, we experiment with reducing the vocabulary size of the XLM-R-base model before adapting via MAFT. There are two possible vocabulary reductions in our setting: \textit{(1) removal of tokens before MAFT or (2) removal of tokens after MAFT}. From our preliminary experiments, we find approach (1) to work better. We call the resulting model, \textit{AfroXLMR-small}. 

To remove non-African vocabulary sub-tokens from the pretrained XLM-base model,  we concatenated the monolingual texts from 19 out of the 20 African languages together. Then, we apply \texttt{sentencepiece} to the Amharic monolingual texts, and concatenated texts separately using the original XLM-R-base tokenizer. The frequency of all the sub-tokens in the two separate monolingual corpora is computed, and we select the top-k most frequent tokens from the separate corpora. We used this separate sampling to ensure that a considerable number of Amharic sub-tokens are captured in the new vocabulary, we justify the choice of this approach in Section \ref{drop}. We assume that the top-k most frequent tokens should be representative of the vocabulary of the whole 20 languages. We chose $k=52.000$ from the Amharic sub-tokens which covers {99.8\%} of the Amharic monolingual texts, and $k=60.000$ which covers {99.6\%} of the other 19 languages, and merged them. In addition, we include the top 1000 tokens from the original XLM-R-base tokenizer in the new vocabulary to include frequent tokens that were not present in the new top-k tokens.\footnote{This introduced just a few new tokens which are mostly English tokens to the new vocabulary. We end up with $70.609$ distinct sub-tokens after combining all of them.} 
We note that our assumption above may not hold in the case of some very distant and low-resourced languages as well as when there are domain differences between the corpora used during adaptation and fine-tuning. We leave the investigation of alternative approaches for vocabulary compression for future work.

\begin{table*}[t]
\footnotesize
 \begin{center}
 \resizebox{\textwidth}{!}{%
  \begin{tabular}{ll|rrrrrrrrrrr|r}
    \toprule
    \textbf{Model} & \textbf{Size} &  \textbf{\texttt{amh}} &\textbf{\texttt{eng}} &\textbf{\texttt{hau}} &  \textbf{\texttt{kin}} & \textbf{\texttt{lin}} & \textbf{\texttt{mlg}} & \textbf{\texttt{pcm}} & \textbf{\texttt{som}} & \textbf{\texttt{swa}} & \textbf{\texttt{yor}} & \textbf{\texttt{zul}} & \textbf{avg} \\
    \midrule
    \multicolumn{4}{l}{\texttt{Finetune}} \\
    XLM-R-miniLM & 117M & 70.4  & 94.1 & 77.6 & 64.2 & 41.2 & 42.9 & 67.6 &  74.2 & 86.7 & 68.8 & 56.9 & 67.7 \\
    AfriBERTa & 126M& 70.7  & 93.6 & 90.1 & 75.8 & 55.4 & 56.4 &  81.5 & \textbf{79.9}  & 87.7 & \textbf{82.6} & 71.4 & 76.8 \\
     \rowcolor{Color}
    XLM-R-base & 270M & 71.1  & 94.1 & 85.9 & 73.3 & 56.8 & 54.2 &  77.3 & 78.8 & 87.1 & 71.1 & 70.0 & 74.6 \\
    XLM-R-large & 550M  & 72.7 & 94.5 & 86.2 & 75.1 & 52.2 & 63.6 & 79.4 & 79.2 & 87.5 & 74.8 & 78.7 & 76.7 \\
    \midrule
    \multicolumn{4}{l}{\texttt{MAFT + Finetune}} \\
    XLM-R-miniLM & 117M& 69.5  & 94.1 & 86.7 & 72.0 & 51.7 & 55.3 & 78.1 &  77.7 & 87.2 & 74.0 & 60.3 & 73.3 \\
    AfriBERTa & 126M & 68.8  & 93.7 & 89.5 & 76.5 & 54.9 & 59.7 &  \textbf{82.2} &  \textbf{79.9} & 87.7 & 80.8 & 76.4 & 77.3\\
     \rowcolor{Color}
    XLM-R-base & 270M & 71.9  & \textbf{94.6} & 88.3 & \textbf{76.8} & \textbf{58.6} & 64.7  &  78.9 &  79.1 & 87.8 & 80.2 & \textbf{79.6} & 78.2 \\
    XLM-R-base-v70k & 140M & 70.4  & 94.2 & 87.7 & 76.1 & 56.8 & 64.4 & 76.1 &  79.4 & 87.4 & 76.9 & 77.4 & 76.9 \\
    \midrule
   \rowcolor{Color}
    XLM-R-base+LAFT & 270M x 11 & \textbf{73.0}  & 94.3 & \textbf{91.2} & 76.0 & 56.9 & \textbf{67.3} &  77.4 & 79.4 & \textbf{88.0} & 79.2 & 79.5 & \textbf{78.4}\\
    \bottomrule
  \end{tabular}
  }
    \caption{\textbf{News topic classification} model comparison, showing F1-score on the test sets after 25 epochs averaged over 5 runs. For LAFT, we multiplied the size of XLM-R-base by the number of languages.}
  \label{tab:news_topic}
  \vspace{-2mm}
  \end{center}
  
\end{table*}

\subsection{Results and discussion}

\begin{table*}[t]
 \begin{center}
 \scalebox{0.90}{
 \footnotesize
  \begin{tabular}{ll|rrrrrr|r}
    \toprule
    \textbf{Model} & \textbf{Size} &   \textbf{\texttt{amh}} & \textbf{\texttt{eng}} & \textbf{\texttt{hau}} & \textbf{\texttt{ibo}} & \textbf{\texttt{pcm}} & \textbf{\texttt{yor}} & \textbf{\texttt{avg}}\\
    \midrule
 
    \multicolumn{2}{l}{\texttt{Finetune}} \\
    XLM-R-miniLM  & 117M & $51.0$ & $62.8$ & $75.0$ & $78.0$ & $72.9$ & $73.4$ & $68.9$ \\
    AfriBERTa-large  & 126M & $51.7$ & $61.8$ & $81.0$ & $\mathbf{81.2}$ & $75.0$ & $80.2$ & $71.8$ \\
    \rowcolor{Color}
    XLM-R-base & 270M & $51.4$ & $66.2$ & $78.4$ & $79.9$ & $76.3$ & $76.9$ & $71.5$ \\
    XLM-R-large & 550M & $52.4$ & $\mathbf{67.5}$ & $79.3$ & $80.8$ & $77.6$ & $78.1$ & 72.6 \\
    \midrule
    \multicolumn{2}{l}{\texttt{MAFT+Finetune}} \\
     XLM-R-miniLM & 117M & $51.3$ & $63.3$ & $77.7$ & $78.0$ & $73.6$ & $74.3$ & $69.7$ \\
     AfriBERTa & 126M & $53.6$ & $63.2$ & $81.0$ & $80.6$ & $74.7$ & $80.4$ & $72.3$ \\
      \rowcolor{Color}
     XLM-R-base & 270M & $53.0$ & $65.6$ & $80.7$ & $80.5$ & $77.5$ & $79.4$ & $72.8$ \\
     XLM-R-base-v70k & 140M & $52.2$ & $65.3$ & $80.6$ & $81.0$ & $77.4$ & $78.6$ & $72.5$ \\
     \midrule
     \rowcolor{Color}
    XLM-R-base+LAFT & 270M x 6 & $\mathbf{55.0}$ & $65.6$ & $\mathbf{81.5}$ & $80.8$ & $74.7$ & $\mathbf{80.9}$ & $\mathbf{73.1}$ \\
    \bottomrule
  \end{tabular}
}
  \caption{\textbf{Sentiment classification} model comparison, showing F1 evaluation on test sets after 20 epochs, averaged over 5 runs. We obtained the results for the baseline model results of ``hau'', ``ibo'', ``pcm'', and ``yor'' from \citet{Muhammad2022NaijaSentiAN}.  For LAFT, we multiplied the size of XLM-R-base by the number of languages as LAFT results in a single model per language.}
  \label{tab:sentiment}
  \end{center}
  \vspace{-3mm}
\end{table*}

\subsubsection{Baseline results}

\begin{table*}[t]
\footnotesize
\begin{center}
 \scalebox{0.85}{
\begin{tabular}{ccc|cc|cc|cc}
\toprule
\textbf{Model}
& \multicolumn{2}{c}{\textbf{\texttt{amh}}} & \multicolumn{2}{c}{\textbf{\texttt{ara}}} & \multicolumn{2}{c}{\textbf{\texttt{eng}}} & \multicolumn{2}{c}{\textbf{\texttt{yor}}} \\
\midrule
& \textbf{\#UNK} & \textbf{F1} & \textbf{\#UNK} & \textbf{F1} & \textbf{\#UNK} & \textbf{F1}  & \textbf{\#UNK} & \textbf{F1} \\
\midrule
AfroXLMR-base & 0 & 76.1 & 0 & 79.7 & 0 & 92.8  & 24 & 82.1 \\
Afro-XLM-R70k (i) & 3704 & 67.8 & 1403 & 76.3 & 44 & 90.6 & 5547 & 81.2 \\
Afro-XLM-R70k (ii) & 3395 & 70.1 & 669 & 76.4 & 54 & 91.0 & 6438 & 81.3 \\
\bottomrule 
\end{tabular} 
}
\footnotesize
  \caption{Numbers of UNKs when the model tokenizers are applied on the NER test sets. }
  \label{tab:unks}
\end{center}
\end{table*}
For the baseline models (top rows in Tables \ref{tab:ner}, \ref{tab:news_topic}, and \ref{tab:sentiment}), we directly fine-tune on each of the downstream tasks in the target language: NER, news topic classification and sentiment analysis. 

\paragraph{Performance on languages seen during pre-training} For NER and sentiment analysis we find XLM-R-large to give the best overall performance. We attribute this to the fact that it has 
a larger model capacity compared to the other PLMs. Similarly, we find AfriBERTa and XLM-R-base to give better results on languages they have been pre-trained on (see \autoref{tab:plm_languages}), and in most cases AfriBERTa tends to perform better than XLM-R-base on languages they are both pre-trained on, for example \texttt{amh}, \texttt{hau}, and \texttt{swa}. However, when the languages are unseen by AfriBERTa (e.g. \texttt{ara}, \texttt{eng},  \texttt{wol}, \texttt{lin}, \texttt{lug}, \texttt{luo}, \texttt{xho}, \texttt{zul}), it performs much worse than XLM-R-base and in some cases even worse than the XLM-R-miniLM. This shows that it may be better to adapt to a new African language from a PLM that has seen numerous languages than one trained on a subset of African languages from scratch. 

\paragraph{LAFT is a strong baseline} The results of applying LAFT to the XLM-R-base model are shown in the last row of Tables \ref{tab:ner}, \ref{tab:news_topic}, and \ref{tab:sentiment}. We find that applying LAFT on each language individually provides a significant improvement in performance across all languages and tasks we evaluated on. Sometimes, the improvement is very large, for example, $+7.4$ F1 on Amharic NER and $+9.5$ F1 for Zulu news-topic classification. The only exception is for English since XLM-R has already seen large amounts of English text during pre-training. Additionally, LAFT models tend to give slightly worse result when adaptation is performed on a smaller corpus.\footnote{We performed LAFT on \texttt{eng} using VOA news corpus with about 906.6MB, much smaller than the CC-100 \texttt{eng} corpus (300GB)} 

\subsubsection{Multilingual adaptive fine-tuning results} 

While LAFT provides an upper bound on downstream performance for most languages, our new approach is often competitive to LAFT. On average, the difference on NER, news topic and sentiment classification is $-0.5$, $-0.2$, and $-0.3$ F1, respectively. Crucially, compared to LAFT, MAFT results in a single adapted model which can be applied to many languages while LAFT results in a new model for each language. Below, we discuss our results in more detail.

\paragraph{PLMs pre-trained on many languages benefit the most from MAFT} We found all the PLMs to improve after we applied MAFT. The improvement is the largest for the XLM-R-miniLM, where the performance improved by $+4.2$ F1 for NER, and $+5.6$ F1 for news topic classification. Although, the improvement was lower for sentiment classification ($+0.8$). Applying MAFT on XLM-R-base gave the overall best result. On average, there is an improvement of $+2.6$, $+3.6$, and $+1.5$ F1 on NER, news topic and sentiment classification, respectively. The main advantage of MAFT is that it allows us to use the same model for many African languages
instead of many models specialized to individual languages. This significantly reduces the required disk space to store the models, without sacrificing performance. Interestingly, there is no strong benefit of applying MAFT to AfriBERTa. In most cases the improvement is $<0.6$ F1. We speculate that this is probably due to AfriBERTa's tokenizer having a limited coverage. We leave a more detailed investigation of this for future work. 

\paragraph{More efficient models using vocabulary reduction} Applying vocabulary reduction helps to reduce the model size by more than $50\%$ before applying MAFT. We find a slight reduction in performance as we remove more vocabulary tokens. Average performance of XLM-R-base-v70k reduces by $-1.6$,  $-1.5$ and $-0.6$ F1 for NER, news topic, and sentiment classification compared to the XLM-R-base+LAFT baseline. 
Despite the reduction in performance compared to XLM-R-base+LAFT, they are still better than XLM-R-miniLM, which has a similar model size, with or without MAFT. We also find that their performance is better than that of the PLMs that have not undergone any adaptation. We find the largest reduction in performance on languages that make use of non-Latin scripts i.e. \texttt{amh} and \texttt{ara} --- they make use of the Ge'ez script and Arabic script respectively. We attribute this to the vocabulary reduction impacting the number of \texttt{amh} and \texttt{ara} subwords covered by our tokenizer. 

In summary, we recommend XLM-R-base+MAFT (i.e. AfroXLMR-base) for all languages on which we evaluated, including high-resource languages like English, French and Arabic. If there are GPU resource constraints, we recommend using XLM-R-base-v70k+MAFT (i.e. AfroXLMR-small). 

\subsection{Ablation experiments on vocabulary reduction}
\label{drop}
Our results showed that applying vocabulary reduction reduced the model size, but we also observed a drop in performance for different languages across the downstream tasks, especially for Amharic, because it uses a non-Latin script. Hence, we compared different sampling strategies for selecting the top-k vocabulary sub-tokens. These include: (i) concatenating the monolingual texts, and selecting the top-70k sub-tokens (ii) the exact approach described in Section \ref{reduce}. The resulting tokenizers from the two approaches are used to tokenize the sentences in the NER test sets for Amharic, Arabic, English, and \yoruba. Table \ref{tab:unks} shows the number of UNKs in the respective test set after tokenization and the F1 scores obtained on the NER task for the languages. 
The table shows that the original AfroXLMR tokenizer obtained the least number of UNKs for all languages, with the highest F1 scores. Note that \yoruba has $24$ UNKs, which is explained by the fact that \yoruba was not seen during pre-training. Furthermore, using approach (i), gave $3704$ UNKs for Amharic, but with approach (ii) there was a significant drop in the number of UNKs and an improvement in F1 score. We noticed a drop in the vocabulary coverage for the other languages as we increased the Amharic sub-tokens. Therefore, we concluded that there is no sweet spot in terms of the way to pick the vocabulary that covers all languages and we believe that this is an exciting area for future work.

\subsection{Scaling MAFT to larger models}

To demonstrate the applicability of MAFT to larger models, we applied MAFT to XLM-R-large using the same training setup as XLM-R-base. We refer to the new PLM as AfroXLMR-large. For comparison, we also trained individual LAFT models using the monolingual data\footnote{For languages not in MasakhaNER, we use the same monolingual data in \autoref{tab:monolingual_corpus}. } from \citet{adelani-etal-2021-masakhaner}. \autoref{tab:ch5_ner_large} shows the evaluation result on NER. Averaging over all 13 languages, AfroXLMR-large improved over XLM-R-large by $+2.8$ F1, which is very comparable to the improvement we obtained between AfroXLMR-base ($81.8$ F1) and XLM-R-base ($79.2$ F1). Surprisingly, the improvement is quite large ($+3.5$ to $+6.3$ F1) for seven out of ten African languages: \texttt{yor}, \texttt{luo}, \texttt{lug}, \texttt{kin}, \texttt{ibo}, and \texttt{amh}. The most interesting observation is that AfroXLMR-large, on average, is either competitive or better than the individual language LAFT models, including languages not seen during the MAFT training stage like \texttt{lug}, \texttt{luo} and \texttt{wol}. This implies that AfroXLMR-large (a single model) provides a better alternative to XLM-R-large+LAFT (for each language) in terms of performance on downstream tasks and disk space. 
AfroXLMR-large is currently the largest masked language model for African languages, and achieves the state-of-the-art compared to all other multilingual PLM on the NER task. This shows that our MAFT approach is very effective and scales to larger PLMs. 

\begin{table*}[t]
\footnotesize
 \begin{center}
  \resizebox{\textwidth}{!}{%
  \begin{tabular}{ll|rrrrrrrrrrrrr|r}
    \toprule
    \textbf{Model} &\textbf{Size} &\textbf{\texttt{amh}} &\textbf{\texttt{ara}} &\textbf{\texttt{eng}} &\textbf{\texttt{hau}}  & \textbf{\texttt{ibo}} & \textbf{\texttt{kin}} & \textbf{\texttt{lug}} & \textbf{\texttt{luo}} & \textbf{\texttt{pcm}} & \textbf{\texttt{swa}} & \textbf{\texttt{wol}} & \textbf{\texttt{xho}} & \textbf{\texttt{yor}} & \textbf{avg} \\
    \midrule
    XLM-R-large & 550M & 76.2 & 79.7 & \textbf{93.1} & 90.5 & 84.1 & 73.8 & 81.6 & 73.6 & 89.0 & 89.4 & 67.9 & 72.4 & 78.9 & 80.8 \\
    XLM-R-large+LAFT & 550M x 13 & \textbf{79.9} & \textbf{81.3} & 92.2 & \textbf{91.7} & \textbf{87.7} & 78.4 & 86.2 & \textbf{78.2} & \textbf{91.1} & 90.3 & 68.8 & 72.7 & 82.9 & 83.2 \\
    AfroXLMR-large  & 550M& 79.7 & 80.9 & 92.2 & 91.2 & \textbf{87.7} & \textbf{79.1} & \textbf{86.7} & 78.1 & 91.0 & \textbf{90.4} & \textbf{69.6} & \textbf{72.9} & \textbf{85.2} & \textbf{83.4}\\
    \bottomrule
  \end{tabular}
  }
    \caption{\textbf{NER} model comparison on XLM-R-large, XLM-R-large+LAFT and XLM-R-large+MAFT (i.e AfroXLMR-large), showing F1-score on the test sets after 50 epochs averaged over 5 runs. Results are for all 4 tags in the dataset: PER, ORG, LOC, DATE/MISC. }
  \label{tab:ch5_ner_large}
  \end{center}
  
\end{table*}

\section{Cross-lingual Transfer Learning} 

\begin{table*}[t]
\footnotesize
 \begin{center}
  \resizebox{\textwidth}{!}{%
  \begin{tabular}{l|rrrrrrrrrr|r}
    \toprule
    \textbf{Model} & \textbf{\texttt{amh}} &\textbf{\texttt{hau}}  & \textbf{\texttt{ibo}} & \textbf{\texttt{kin}} & \textbf{\texttt{lug}} & \textbf{\texttt{luo}} & \textbf{\texttt{pcm}} & \textbf{\texttt{swa}} & \textbf{\texttt{wol}} & \textbf{\texttt{yor}} & \textbf{avg} \\
    \midrule
    XLM-R-base (fully-supervised) & \textbf{69.7} & \textbf{91.0} & \textbf{86.2} & \textbf{73.8} & 80.5 & \textbf{75.8} & \textbf{86.9} & \textbf{88.7} & 69.6 & 78.1 & 81.2\\
    \midrule
    mBERT (MAD-X)~\cite{ansell2021composable} & - & 83.4 & 71.7 & 65.3 & 67.0 & 52.2 & 72.1 & 77.6 & 65.6 &  74.0 & 69.9 \\
    mBERT (MAD-X on \textbf{news domain}) & - & 86.0 & 77.6 & 69.9  & 73.3 & 56.9 & 78.5 & 80.2 & 68.8 & 75.6 & 74.1 \\
    XLM-R-base (MAD-X on \textbf{news domain}) & 47.5 & 85.5 & 83.2 & 72.0  & 75.7 & 57.8 & 76.8 & 84.0 & 68.2 & 72.2 & 75.0 \\
    AfroXLMR-base (MAD-X on \textbf{news domain}) & 47.7 & 88.1 & 80.9 & 73.0  & 80.1 & 59.2 & 79.9 & \underline{86.9} & 69.1 & 75.6 & 77.0  \\
    \midrule
    mBERT (LT-SFT)~\cite{ansell2021composable} & - & 83.5 & 76.7 & 67.4 & 67.9 & 54.7 & 74.6 & 79.4 & 66.3 &  74.8 & 71.7 \\
    mBERT (LT-SFT on \textbf{news domain}) & - & 86.4 & 80.6 & 69.2 & 76.8 & 55.1 & 80.4 & 82.3 & 71.6 & 76.7 & 75.4 \\
   
    XLM-R-base (LT-SFT on \textbf{news domain}) & \underline{54.1} & 87.6 & 81.4 & 72.7 & 79.5 & \underline{60.7} & 81.2 & 85.5 & 73.6 & 73.7 & 77.3\\
    AfroXLMR-base (LT-SFT on \textbf{news domain}) & 54.0 & \underline{88.6} & \underline{83.5} & \textbf{\underline{73.8}} & \textbf{\underline{81.0}} &\underline{60.7} & \underline{81.7} & 86.4 & \textbf{\underline{74.5}} & \textbf{\underline{78.7}} & \underline{78.8}\\
   
    \bottomrule
  \end{tabular}
  }
    \caption{\textbf{Cross-lingual transfer using LT-SFT}~\cite{ansell2021composable} and evaluation on \textbf{MasakhaNER}. The full-supervised baselines are obtained from \citet{adelani-etal-2021-masakhaner} to measure performance gap when annotated datasets are available. Experiments are performed on 3 tags: PER, ORG, LOC. Average (avg) excludes \texttt{amh}. The best zero-shot transfer F1-scores are \underline{underlined}.}
  \label{tab:lt_sft}
  \vspace{-3mm}
  \end{center}
  
\end{table*}


The previous section demonstrates the applicability of MAFT in the fully-supervised transfer learning setting. Here, we demonstrate that our MAFT approach is also very effective in the zero-shot cross-lingual transfer setting using parameter-efficient fine-tuning methods.

Parameter-efficient fine-tuning methods like \textit{adapters}~\cite{pmlr-v97-houlsby19a} are appealing because of their modularity, portability, and composability across languages and tasks. Often times, language adapters are trained on a general domain corpus like Wikipedia. However, when there is a mismatch between the target \textit{domain} of the task and the \textit{domain} of the language adapter, it could also impact the cross-lingual performance. 

Here, we investigate how we can improve the cross-lingual transfer abilities of our adapted PLM -- AfroXLMR-base by training language adapters on the same domain as the target task. For our experiments, we use the MasakhaNER dataset, which is based on the news domain. We compare the performance of language adapters trained on \textit{Wikipedia} and \textit{news} domains. In addition to adapters, we experiment with another parameter-efficient method based on Lottery-Ticket Hypothesis~\cite{frankle2018the} i.e. LT-SFT~\cite{ansell2021composable}. 

For the adapter approach, we make use of the MAD-X approach~\cite{pfeiffer-etal-2020-mad} -- an adapter-based framework that enables cross-lingual transfer to arbitrary languages by learning modular language and task representations. However, the evaluation data in the target languages should have the same task and label configuration as the source language. Specifically, we make use of MAD-X 2.0~\cite{pfeiffer-etal-2021-unks}  where the last adapter layers are dropped, which has been shown to improve performance. The setup is as follows: (1) We train language adapters via masked language modelling (MLM) individually on source and target languages, the corpora used are described in Appendix A.2; (2) We train a task adapter by fine-tuning on the target task using labelled data in a source language. (3) During inference, task and language adapters are stacked together by substituting the source language adapter with a target language adapter. 

We also make use of the Lottery Ticket Sparse Fine-tuning (LT-SFT) approach~\citep{ansell2021composable}, a parameter-efficient fine-tuning approach that has been shown to give competitive or better performance than the MAD-X 2.0 approach.  The LT-SFT approach is based on the Lottery Ticket Hypothesis (LTH) that states that each
neural model contains a sub-network (a ``winning ticket'') that, if trained again in isolation, can reach or even surpass the performance of the original model. The LTH is originally a compression approach, the authors of LT-SFT re-purposed the approach for cross-lingual adaptation by finding sparse sub-networks for tasks and languages, that will later be composed together for zero-shot adaptation, similar to Adapters. For additonal details we refer to \citet{ansell2021composable}. 

\subsection{Experimental setup}
For our experiments, we followed the same setting as \citet{ansell2021composable} that adapted mBERT from English CoNLL03~\cite{tjong-kim-sang-de-meulder-2003-introduction} to African languages (using MasakhaNER dataset) for the NER task.\footnote{We excluded the MISC and DATE from CoNLL03 and MasakhaNER respectively to ensure same label configuration.} Furthermore, we extend the experiments to XLMR-base and AfroXLMR-base. For the training of MAD-X 2.0 and sparse fine-tunings (SFT) for African languages, we make use of the monolingual texts from the news domain since it matches the domain of the evaluation data. Unlike, \citet{ansell2021composable} that trained adapters and SFT on monolingual data from Wikipedia domain except for \texttt{luo} and \texttt{pcm} where the dataset is absent, we show that the domain used for training language SFT is also very important. For a fair comparison, we reproduced the result of \citet{ansell2021composable} by training MAD-X 2.0 and LT-SFT on mBERT, XLM-R-base and AfroXLMR-base on target languages with the news domain corpus. But, we still make use of the pre-trained English language adapter\footnote{\url{https://adapterhub.ml/}} and  SFT\footnote{\url{https://huggingface.co/cambridgeltl}} for mBERT and XLM-R-base trained on the Wikipedia domain. For the AfroXLMR-base, we make use of the same English adapter and SFT as XLM-R-base because the PLM is already good for English language. We make use of the same hyper-parameters reported in the LT-SFT paper. 

\paragraph{Hyper-parameters for adapters} We train the task adapter using the following hyper-parameters: batch size of 8, 10 epochs, ``pfeiffer'' adapter config, adapter reduction factor of 8, and learning rate of 5e-5. For the language adapters, we make use of 100 epochs or maximum steps of 100K, minimum number of steps is 30K, batch size of 8, ``pfeiffer+inv'' adapter config, adapter reduction factor of 2, learning rate of 5e-5, and maximum sequence length of 256. For a fair comparison with adapter models trained on Wikipedia domain, we used the same hyper-parameter settings~\cite{ansell2021composable} for the news domain.

\subsection{Results and discussion}

\autoref{tab:lt_sft} shows the results of MAD-X 2.0 and LT-SFT, we compare their performance to fully supervised setting, where we fine-tune XLM-R-base on the training dataset of each of the languages, and evaluate on the test-set. We find that both MAD-X 2.0 and LT-SFT using news domain for African languages produce better performance ($+4.2$ on MAD-X and $+3.7$ on LT-SFT) than the ones trained largely on the wikipedia domain. This shows that the domain of the data matters. Also, we find that training LT-SFT on XLM-R-base gives better performance than mBERT on all languages. For MAD-X, there are a few exceptions like \texttt{hau}, \texttt{pcm}, and \texttt{yor}. Overall, the best performance is obtained by training LT-SFT on AfroXLMR-base, and sometimes it give better performance than the fully-supervised setting (e.g. as observed in \texttt{kin} and \texttt{lug}, \texttt{wol} \texttt{yor} languages). On both MAD-X and LT-SFT, AfroXLMR-base gives the best result since it has been firstly adapted on several African languages and secondly on the target domain of the target task. This shows that the MAFT approach is effective since the technique provides a better PLM that parameter-efficient methods can benefit from. 


\section{Conclusion}

In this work, we proposed and studied MAFT as an approach to adapt multilingual PLMs to many African languages with a single model. We evaluated our approach on 3 different NLP downstream tasks and additionally contribute novel news topic classification dataset for 4 African languages. Our results show that MAFT is competitive to LAFT while providing a single model compared to many models specialized for individual languages. We  went further to show that combining vocabulary reduction and MAFT leads to a {50\%} reduction in the parameter size of a XLM-R while still being competitive to applying LAFT on individual languages. We hope that future work improves vocabulary reduction to provide even smaller models with strong performance on distant and low-resource languages. To further research on NLP for African languages and reproducibility, we have uploaded our language adapters, language SFTs, AfroXLMR-base, AfroXLMR-small, and AfroXLMR-mini models to the HuggingFace Model Hub\footnote{\url{https://huggingface.co/models?sort=downloads&search=Davlan\%2Fafro-xlmr}}.

\section*{Acknowledgments}
Jesujoba Alabi was partially funded by the BMBF project SLIK under the Federal Ministry of Education and Research grant 01IS22015C. David Adelani acknowledges the EU-funded Horizon 2020 projects:  ROXANNE under grant number 833635 and COMPRISE (\texttt{http://www.compriseh2020.eu/}) under grant agreement No. 3081705. Marius Mosbach acknowledges funding from the Deutsche Forschungsgemeinschaft (DFG, German Research Foundation) – Project-ID 232722074 – SFB 1102. We also thank DFKI GmbH for providing the infrastructure to run some of the experiments. 
We are grateful to CoreWeave and EleutherAI for providing the compute to train AfroXLMR-large. 
We thank Alan Ansell for providing his MAD-X 2.0 code. Lastly, we thank Benjamin Muller, the anonymous reviewers of AfricaNLP 2022 workshop and COLING 2022 for their helpful feedback.

\bibliography{iclr2022_conference}
\bibliographystyle{iclr2022_conference}

\appendix
\section{Appendix}

\begin{table*}[ht]
\footnotesize
 \begin{center}
\resizebox{\textwidth}{!}{%
\begin{tabular}{llrr}

\toprule
\textbf{ Language} & \textbf{Source} & \textbf{Size (MB)} & \textbf{No. of sentences}  \\
\midrule
Afrikaans (afr) & mC4 (subset)~\citep{xue-etal-2021-mt5} & 752.2MB & 3,697,430\\
Amharic (amh) & mC4 (subset), and VOA & 1,300MB & 2,913,801\\
Arabic (ara) & mC4 (subset) & 1,300MB & 3,939,375\\
English (eng) & mC4 (subset), and VOA & 2,200MB & 8,626,571 \\
French (fra) & mC4 (subset), and VOA & 960MB & 4,731,196 \\
Hausa (hau) & mC4 (all), and VOA & 594.1MB & 3,290,382\\
Igbo (ibo) & mC4 (all), and AfriBERTa Corpus~\citep{ogueji-etal-2021-small} & 287.5MB & 1,534,825\\
Malagasy (mlg) & mC4 (all) & 639.6MB & 3,304,459 \\
Chichewa (nya) & mC4 (all), Chichewa News Corpus~\citep{Siminyu2021AI4DA} & 373.8MB & 2,203,040 \\
Oromo (orm) & AfriBERTa Corpus, and VOA & 67.3MB & 490,399\\
Naija (pcm) & AfriBERTa Corpus & 54.8MB & 166,842\\
Rwanda-Rundi (kin/run) & AfriBERTa Corpus, KINNEWS \& KIRNEWS~\citep{niyongabo-etal-2020-kinnews}, and VOA & 84MB & 303,838\\
chiShona (sna) & mC4 (all), and VOA & 545.2MB & 2,693,028\\
Somali (som) & mC4 (all), and VOA & 1,000MB & 3,480,960 \\
Sesotho (sot) & mC4 (all) & 234MB & 1,107,565 \\
Kiswahili (swa) & mC4 (all) & 823.5MB & 4,220,346 \\
isiXhosa (xho) & mC4 (all), and Isolezwe Newspaper  & 178.4MB &  832,954 \\
\yoruba (yor) & mC4 (all), Alaroye News, Asejere News, Awikonko News, BBC, and VON  & 179.3MB & 897,299 \\
\zulu (zul) & mC4 (all), and Isolezwe Newspaper & 700.7MB & 3,252,035  \\
\bottomrule 

\end{tabular}
}
\footnotesize
  \caption{Monolingual Corpora (after pre-processing -- we followed AfriBERTa~\cite{ogueji-etal-2021-small} approach) , their sources and size (MB), and number of sentences. }
  \label{tab:monolingual_corpus}
\end{center}
\end{table*}

\begin{table*}[ht]
\footnotesize
 \begin{center}
\resizebox{\textwidth}{!}{%
\begin{tabular}{llrr}

\toprule
\textbf{ Language} & \textbf{Source} & \textbf{Size (MB)} & \textbf{No. of sentences}  \\
\midrule
Amharic (amh) & VOA~\cite{DBLP:journals/corr/abs-2201-05609} & 19.9MB & 72,125\\
Hausa (hau) & VOA~\cite{DBLP:journals/corr/abs-2201-05609} & 46.1MB & 235,614\\
Igbo (ibo) & BBC Igbo~\citep{ogueji-etal-2021-small} & 16.6MB & 62,654\\
Kinyarwanda (kin) & KINNEWS~\cite{niyongabo-etal-2020-kinnews} & 35.8MB & 61,910\\
Luganda (lug) & Bukedde & 7.9MB & 67,716 \\
Luo (luo) & Ramogi FM news and MAFAND-MT~\cite{adelani-etal-2022-thousand} & 1.4MB & 8,684 \\
Naija (pcm) & BBC & 50.2MB & 161,843\\
Kiswahili (swa) & VOA~\cite{DBLP:journals/corr/abs-2201-05609} & 17.1MB & 88,314\\
Wolof (wol) & Lu Defu Waxu, Saabal, Wolof Online, and MAFAND-MT~\cite{adelani-etal-2022-thousand}  & 2.3MB &  13,868 \\
\yoruba (yor) & BBC \yoruba  & 15.0MB & 117,124 \\
\bottomrule 

\end{tabular}
}
\footnotesize
  \caption{Monolingual News Corpora used for language adapter and SFT training, their sources and size (MB), and number of sentences. }
  \label{tab:news_corpus}
\end{center}
\end{table*}

\subsection{Monolingual corpora for LAFT and MAFT}
\label{appendix:maft-corpora}
For training the MAFT models, we make use of the aggregation of monolingual data from \autoref{tab:monolingual_corpus}. 

For the LAFT models, we make use of existing XLMR-base+LAFT models from the MasakhaNER paper~\citep{adelani-etal-2021-masakhaner}. However, for other languages not present in MasakhaNER (\texttt{ara}, \texttt{mlg},\texttt{orm}, \texttt{sna}, \texttt{som}, \texttt{xho}), we make use of the mC4 corpus except for \texttt{eng} --- we use the VOA corpus. For a fair comparison across models, when training the XLM-R-large+LAFT models, we use the same monolingual corpus used to train XLM-R-base+LAFT models. 

\subsection{News corpora for language adapters and SFTs}
\label{appendix:news-corpora-adapters}
\autoref{tab:news_corpus} provides the news corpus we used to train language adapters and SFTs for the cross-lingual settings. 

\end{document}